%% file: main.tex
\documentclass{Interspeech2024}

\interspeechcameraready

\title{Personalized Clinical Note Generation from Doctor-Patient Conversations}

\name[affiliation={1}]{Nathan}{Brake}
\name[affiliation={1}]{Thomas}{Schaaf}

\address{
  $^1$Solventum Health Information Systems, USA
    }
\email{nbrake,tschaaf@solventum.com}

\input{keywords}

\usepackage{algorithm,algpseudocode}
\begin{document}

\maketitle

\input{abstract}

\section{Introduction}
\label{sec:intro}

In the medical domain, physicians are responsible for completing documentation that summarizes key aspects of a patient's visit, for both historical and billing purposes. The task of generating a clinical note is similar to the task of abstractive summarization, since the clinical documentation is often a brief summary of what took place during the patient visit, including but not limited to the reason and background of the patient visit ("History of Present Illness"), any examination performed by the doctor ("Physical Examination"), and any actions that should be taken based on the physician's evaluation ("Assessment \& Plan"). 

Although the general format of these clinical notes may be shared among different physicians, specific phrases and formatting of these notes may vary. For example, one hospital system may require a note to include the use of the appropriate gendered pronouns, while another may require the use of "the patient" instead of using a pronoun. Additionally, a physician may have particular tendencies in the way they hold a conversation with a patient, such as usually beginning with irrelevant small talk, or using casual language to refer a specific clinical term that should appear in the note. When training a model to generate a clinical note given a conversation as input, the model may implicitly learn the formatting of notes based on information it is able to isolate during training. However, when tasked with the creation of a clinical note for a new physician, the model does not have an implicit understanding of this physician's preference or habits. In order to achieve optimal performance for a new physician being enrolled in the system, it is necessary to develop a technique for leveraging their existing data.

\section{Related Work}
\label{sec:reference_work}

Personalization has been a long studied topic in the field of Automatic Speech Recognition (ASR) in order to reduce Word Error Rate (WER). Various approaches for low resource speaker adaptation include the use of I-Vectors \cite{6707705} and data augmentation with X-Vectors \cite{8461375}. Recent efforts have also focused on the effect of using personalized text-to-speech synthesis to boost performance of speakers, especially those who are underrepresented in the available data \cite{10096971}.

In the field of text summarization, the topic of personalization has been studied in the context of review summarization \cite{Li_Li_Zong_2019}. Their research focuses on the automatic generation of a personalized one sentence summary of a short review (average length of 155 words) using an LSTM based Sequence-to-Sequence network. Their work focuses on short reviews and summaries and involves separate user and user-vocabulary embeddings.

The HydraSum \cite{goyal-etal-2022-hydrasum} approach incorporated a multi-decoder training objective which automatically learns contrasting summary and styles. 
The approach focusing on a broader level of style using a small number of parameters to control for example attractiveness or different levels of specificity, but is a text resembles a certain author.

The task of summarizing clinical notes has been explored in the context of generating a summary directly from the doctor-patient conversations (DoPaCos) \cite{Zhang2021LeveragingPM}, as well as the use of Large Language Models such as GPT-3 to summarize the conversation \cite{Nair2023GeneratingMS}. Summarizing a DoPaCo is related to meeting audio summarization with the AMI corpus \cite{7529878bc1a143dbad4fa019e742fdb8}, but differs in that the output summary must be customized based on the author and uses a different vocabulary from that of the original conversation. Additionally, work has explored the value of detecting whether or not a physician is directly dictating to a scribe or software system, and how individual physicians exhibit consistent patterns when they perform these dictations in their conversations  \cite{9688118}. The MediQA \cite{ben-abacha-etal-2023-overview} task uses the ACI-Bench \cite{yim2023acibench} dataset for clinical note summarization, which does not contain any author information for training or evaluation physician specific conversation/note styles.

\section{Data}
\label{sec:data}

For our experiments, we use an internal dataset containing audio recordings of 24,688 DoPaCos with accompanying clinical notes that were completed by human scribes. All conversations are from the Orthopedic specialty, and the audio recordings of these DoPaCos range in length from 4 minutes to 45 minutes. The audio recordings are converted to text using a speech recognition system, designed for recognition of medical conversations. The ASR system was configured to insert punctuation but not to perform speaker diarization, meaning that the output is a paragraph of text without speaker identification tags. Because no manual transcripts exist for this dataset the exact Word Error Rate (WER) of the ASR system is unknown. However, the performance of the ASR system was measured using a similar internal medical conversation dataset of 23 different physicians with 5 different DoPaCos each. The performance on that dataset is calculated via \textsc{sclite v2.10} from the \textsc{SCTK} toolkit\footnote{https://github.com/usnistgov/SCTK} and summarized in Table \ref{wer}. These relatively high WER values are consistent with the measured results of other systems tested on ambient clinical audio in \cite{wer_comparison}. These higher WER are likely related to the inherent difficulty of multi-party ambient audio transcription as opposed to direct dictation.

\begin{table}[htpb]
\centering
\begin{tabular}{llll}
\hline
Spk & \#Sent/Words & Corr/Sub/Del/Ins & WER\\
\hline
DOC & 8,611/64,280 & 74.5/11.4/14.1/5.9 & 31.4 \\ 
PAT & 5,375/27,676 & 58.8/14.1/27.1/6.2 & 47.3 \\ 
\end{tabular}
\caption{
WER of Representative Medical Conversation Dataset
}\label{wer}
\end{table}

The DoPaCos in our dataset come from an asynchronous scribing configuration, meaning that the physician is aware the audio is being provided to a human scribe after the encounter. Because of this, the audio may contain portions where the doctor is directly instructing the scribe what information should be included in the note. For example, a doctor may dictate "The patient's blood pressure was 120 over 80". The doctor may also say something intended to trigger the insertion of some standard phrasing, for example: "Three month populate Smith check and see how she's doing". This statement indicates to the Scribe that the summary should include the sentence "The patient will follow up with John Smith, NP-C, in 3 months to review her progress". Although an AI model could be explicitly taught these shortcut phrases, it may be preferable to enable a model to implicitly learn these relationships based on existing data from a physician.

The History of Present Illness (HPI) section is written in a narrative form and describes the reason for the patient's visit and some description of the relevant medical history. This section may contain information that the doctor directly dictates before and/or after the patient is in the room (e.g. "Jane is a pleasant 50 year old female"), and also information that the patient mentions during the conversation (e.g. "Hi doctor, my knee has been really bothering me after I fell last week").

The Assessment \& Plan (A\&P) and Physical Examination (PE), in contrast to HPI, may often include information that was never directly dictated or mentioned, but rather is inserted based on doctor direction or from an EHR system. For instance, a phrase such as "patient will return after injection" or "add cortisone injection follow up" may indicate that the note should contain a standard wording regarding who and when the patient should return for a follow-up appointment. Content such as exam or diagnostic results may also be present although not mentioned in the conversation. 

From this dataset, we create a train validation, and evaluation split, where all physicians in validation and evaluation are represented in the training split.

We additionally create two splits for exploring the extension of personalized notes for new physicians. First, an adaptation split which contains 10 physicians not seen in the training split. There are 20 conversations for each of these 10 previously unseen physicians, for a total of 200 conversations in the adaptation split. 6 of these 10 physicians are from the same hospital as at least one physician in the training set, while 4 of the 10 physicians are from hospitals not seen in the training set.

Lastly, a holdout physicians test-adapt split is created, which contains a total of 706 conversations from the 10 unique physicians that were in the adaptation split. These 10 physicians have between 32 and 98 encounters. Table \ref{dataset-info} presents a summary of the dataset characteristics.

\begin{table*}[htpb]
\centering
\begin{tabular}{llllllll}
\hline
Dataset & \#conv & \#tok & audio(hr) & \#doc/\#hosp  & conv avg length (min) & HPI avg words & A\&P avg words\\
\hline
train & 21k & 21.5M & 3,145 & 62 / 27 & 1427 & 117 & 134 \\
validation & 1.3k & 1.4M & 200 & 45 / 20 &  1422 &  115 & 131 \\
evaluation & 1.4k & 1.5M & 213 & 46 / 21 &  1430 &  117 & 119 \\
\hline
adapt & 200 & 204k & 22 & 10 / 9 & 981 & 158 & 136 \\
test-adapt & 706 & 723k & 83 & 10 / 9 & 1030 & 152 & 132  \\
\end{tabular}
\caption{\label{dataset-info}
Dataset Split Information 
}
\end{table*}

\section{Method}

Our experiment consists of three phases: training, adaptation, and testing. The training phase trains models of several configurations with the training dataset for clinical note generation. The purpose of the adaptation phase is to execute Algorithm \ref{alg:cap} in order to determine what existing physician embedding should be used for each new physician. This allows for the initialization of a new embedding for the new physician using the embedding of the existing physician. The testing phase then evaluates the success of the adaptation phase and generates statistics that can be used to compare the performance against a baseline model that does not use a physician embedding.

\subsection{Training}
For training, we use the transformer \cite{NIPS2017_3f5ee243} based PEGASUS-X model \cite{phang2022investigating}, which builds on top of the PEGASUS model \cite{pmlr-v119-zhang20ae} and uses a modified attention mechanism in order to support longer input sequences of up to 16,384 tokens, and has been additionally pre-trained using long sequence data. We expect that our method will also be successful when used with other transformer based models. We train a single model to output a single section, one model to output an HPI section, another model to output an A\&P section, and one model to output a PE section.

For each of the 62 unique physicians seen in training, we create a new token in the tokenizer and model vocabulary, which is assigned a trainable embedding (The embedding is represented by a vector of length 768). 

In order to explore where the physician embedding will have the greatest impact, we train 3 different model configurations: prefixing the token to the ASR transcript on the encoder side of the model (ENC), prefixing to the target note section on the decoder side (DEC) , and prefixing to both encoder and decoder side (ENC+DEC). We also train a baseline model (BASE) which does not use any physician token. This results in a total of 12 models trained for this experiment, 1 set of 4 models for each of the 3 clinical note sections.


\subsection{Adaptation}
We run the adaptation phase for the ENC, DEC, and ENC+DEC model designs. Because BASE does not have a trained physician embedding, the adaptation phase is not applicable to that configuration. For each of the 10 physicians in the adaptation dataset, we perform Algorithm \ref{alg:cap}, where \textit{train\_phys\_emb} are the 62 existing physician trained embeddings, \textit{new\_phys\_docs} are the transcript and note pairs available for the new physician, \textit{generate\_note()} is a function that generates a hypothesis clinical note section using the \textit{physician\_embedding}, \textit{score()} is a function that performs a ROUGE-2 score between the hypothesis and reference note, and \textit{num\_docs} is the number of \textit{new\_phys\_docs} being used for adaptation. The algorithm returns \textit{best\_emb}, which is the existing physician embedding that achieves the best ROUGE-2 score for the new physician. Although we use ROUGE-2, this algorithm can easily be adapted for use with another metric. In future work we plan to explore whether additional performance gains could be realized through the training of the selected initial physician embedding using the initial transcript-note pairs from a new physician.

\begin{algorithm}
    \caption{Adaptation Algorithm for Physician}
    \label{alg:cap}
    \begin{algorithmic}[1]
        \State $train\_phys\_emb \gets author\_0, ... , author\_61$
        \State $best\_emb \gets None$
        \State $best\_score \gets 0$
        \For{$phys\_emb$ \text{ in } $train\_phys\_embs$}
            \State $avg\_score \gets 0$
            \For{$transcript$ , $ref\_note$ \text{ in } $new\_phys\_docs$}
                \State $hypo\_note \gets \text{generate\_note}(phys\_emb, transcript)$
                \State $avg\_score \mathrel{+}= \text{score}(hypo\_note, ref\_note)$
            \EndFor
            \State $avg\_score \gets \frac{avg\_score}{\text{num\_docs}}$
            \If{$avg\_score > best\_score$}
                \State $best\_emb \gets phys\_emb$
                \State $best\_score \gets avg\_score$
            \EndIf
        \EndFor

        \State \textbf{return} $best\_emb$
    \end{algorithmic}
\end{algorithm}

\subsection{Testing \& Evaluation}

We use ROUGE \cite{lin-2004-rouge} and Factuality \cite{glover-etal-2022-revisiting} for evaluation of note quality. ROUGE is intended to be a recall based method which helps to encapsulate the transfer of phrasing and format. The Factuality method is intended to evaluate the truthfulness of the generated note. The Factuality model used is trained with proprietary medical data to support clinical note scoring. ROUGE-2 was utilized for the adaptation phase, with the expectation that ROUGE-2 is useful to indicate how well stylistic and conceptual elements of a clinical note are captured.

The adaptation step uses the 20 transcript-note pairs for each new physician to determine which existing physician embedding is the best match for the new physician and will be used in the testing phase. Note that because each section has a different model and is adapted separately, a new physician may have 3 different physician embeddings, one embedding for use with the HPI model, one for the PE model, and one for the A\&P model.

Using the physician ID embeddings that were determined via the adaptation phase, we test the performance of these selections on a test-adapt set of 706 transcripts from the new physicians, to see if the performance of these models are able to improve on the BASE model design that does not use a physician embedding

\section{Results} 

\subsection{Evaluation on Seen Authors}
Table \ref{train-results} presents the results of the models scored with the evaluation split of the dataset, which are physicians that exist in the training split. The models trained with the use of physician embeddings outperform the BASE model. The models that utilize the physician embedding on the decoder tend to have the highest performance.

\begin{table}[htpb]
\centering
\begin{tabular}{lllll}
\hline
\textbf{Model} & \textbf{R-1} & \textbf{R-2} & \textbf{R-L} &  \textbf{Factuality} \\
\hline
HPI BASE & 57.53 & 36.77 & 47.41 &  59.53 \\
HPI ENC & 57.77 & 37.26 & 47.84 &  60.15 \\
HPI DEC & \textbf{59.00} & \textbf{38.66} & \textbf{48.84} & 61.61 \\
HPI ENC+DEC & 58.61 & 38.17 & 48.52 & \textbf{61.64} \\
\hline
A\&P BASE & 51.52 & 36.35 & 44.39 &  66.34 \\
A\&P ENC & \textbf{55.62} & 42.37  & 49.76 &  \textbf{70.78} \\
A\&P DEC & 54.50 & 40.42 & 47.88 &  69.95 \\
A\&P ENC+DEC & 55.21 & \textbf{42.66} & \textbf{49.86} &  70.55 \\
\hline
PE BASE & 56.47 & 45.34 & 51.47 &  69.31 \\
PE ENC & 58.94 & 51.10 & 55.75 &  76.28 \\
PE DEC & \textbf{60.11} & \textbf{53.13} & \textbf{57.45} &  77.67 \\
PE ENC+DEC & 59.87 & 52.87 & 57.16 & \textbf{78.06} \\
\end{tabular}
\caption{\label{train-results}
Rouge (R) and Factuality Metrics for models scored against the evaluation split of known authors.
}
\end{table}

\subsection{Model Adaptation for Unseen Authors}
We observe whether the physician selected by the adaptation process is affected by how many transcripts are used in the adaptation process. For our main results we use 20 transcript-note pairs in the adaption process. Table \ref{adapt-hospital} presents how often the selected physician embedding changes when the adaptation phase considers fewer than the 20 transcript-note pairs available. The lower the number, the fewer physician embeddings were changed, which reflects a more stable adaptation.

\begin{table}[htpb]
\centering
\begin{tabular}{lllllll}
\hline
\textbf{Model} & \textbf{1}  & \textbf{5} & \textbf{10} & \textbf{15} \\
\hline
HPI ENC    &  10 & 7 & 5 & 4 \\
HPI DEC    &  10 & 9 & 6 & 4 \\
HPI ENC+DEC & 9 & 7  & 3 & 0 \\
\hline
A\&P ENC     & 5 & 5 & 5 & 2\\
A\&P DEC     & 9 & 6 & 3 & 3\\
A\&P ENC+DEC & 6 & 4 & 2 & 1\\
\hline
PE ENC      & 6 & 3 & 2 & 1\\
PE DEC      & 8 & 6 & 4 & 2\\
PE ENC+DEC  & 9 & 5 & 4 & 3 \\
\end{tabular}
\caption{\label{adapt-hospital}
The columns indicate the number of transcript-note pairs used in adaptation, and the rows represent how many of the 10 new physicians were assigned a different physician embedding than when the full 20 transcript-note pairs are used.
}
\end{table}

\subsection{Clustering of trained Author Embeddings }

\begin{figure}[htpb]
\centering
\includegraphics[width=0.72\linewidth]{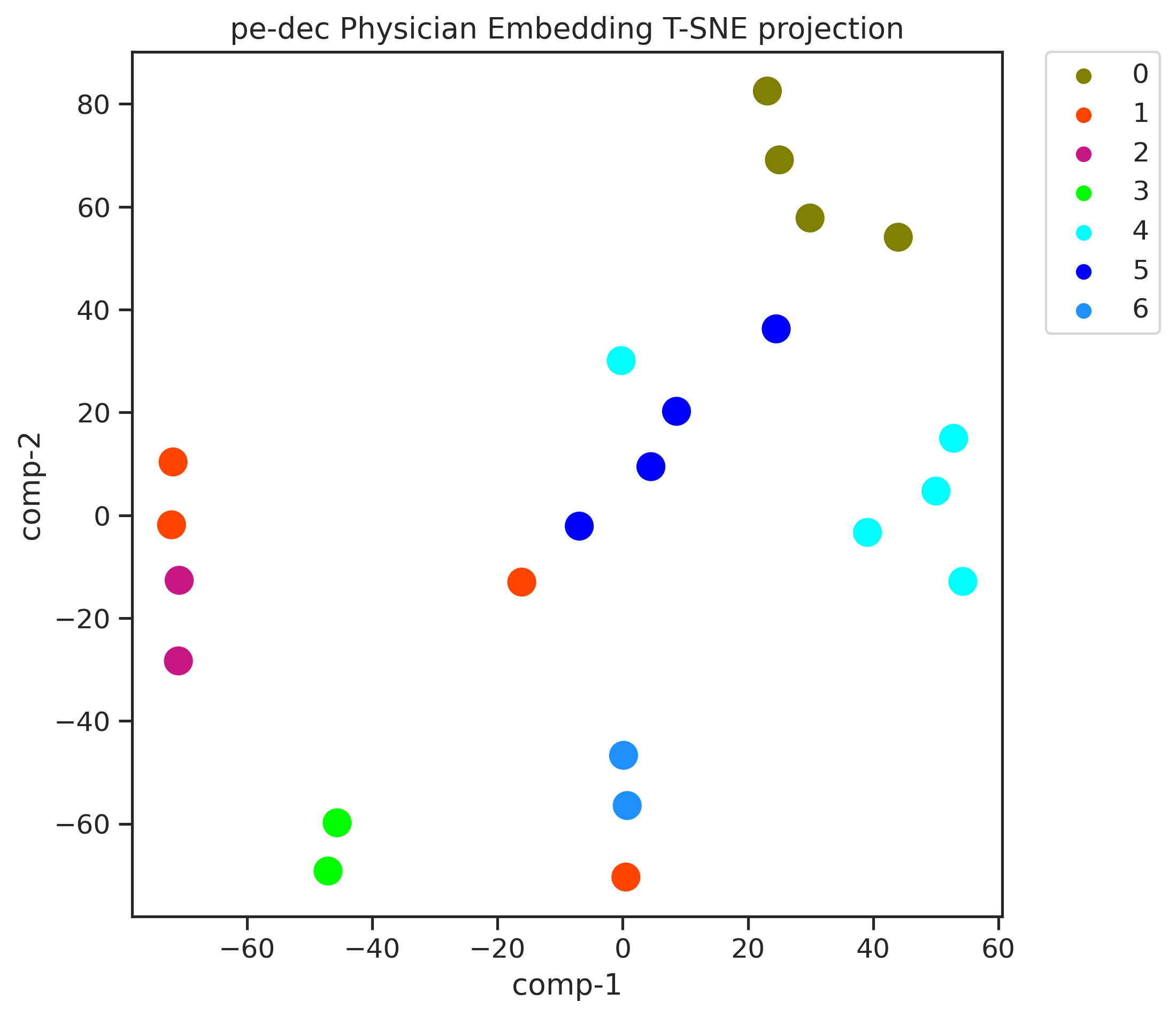} \\
\caption{t-SNE projection for DEC embeddings}\label{fig:tsne}
\end{figure}

In order to better visualize the trained physician embeddings, we use t-SNE to reduce the 768 dimensional embedding vectors into 2D representations. In Figure \ref{fig:tsne}, each color signifies a hospital system, and each point represents a physician embedding. The t-SNE chart illustrates that many hospitals exhibit identifiable clustering of physicians from their hospital.

\subsection{Model Performance after Adaptation}

Our results from the testing phase after adaptation are presented in Table \ref{test-results}. We find that in general, our best adapted models have the physician embedding applied to the decoder, and especially in the A\&P section there is also a benefit to applying the physician embedding to the encoder. An interesting result is the $\sim$15 ROUGE-2 and Factuality improvement for the PE and A\&P sections between the baseline models and the special token prefixed models. Since the PE and A\&P sections can contain a content not specifically dictated and stylistic elements (i.e. the phrasing "Patient is alert and oriented x3" which may have never been explicitly mentioned in the transcript), the adaptation stage is helping to match the new physician with these elements that aren't able to be determined from a transcript alone.

\begin{table}[htpb]
\centering
\begin{tabular}{llllc}
\hline
\textbf{Model} & \textbf{R-1} & \textbf{R-2} & \textbf{R-L} & \textbf{Factuality} \\
\hline
HPI BASE & 49.86 & 27.09 &37.05 &  57.84 \\
HPI ENC & 52.64 & 29.78 &39.95 &  58.82 \\
HPI DEC & \textbf{53.39} & \textbf{30.84} & 40.34 &  60.29 \\
HPI ENC+DEC & 53.35 & 30.47 & \textbf{40.52} & \textbf{61.00} \\
\hline
A\&P BASE & 43.12 & 25.08 & 33.49 &  57.68 \\
A\&P ENC & 52.37 & 35.97 &44.24 & \textbf{70.77} \\
A\&P DEC & 51.78 & 33.87 &42.72 & 69.27 \\
A\&P ENC+DEC & \textbf{54.52} & \textbf{37.83} & \textbf{46.26} & 70.03 \\
\hline
PE BASE & 33.74 & 16.78 &25.18 & 48.10 \\
PE ENC & 44.22 & 29.17 &36.60 &  60.15 \\
PE DEC & \textbf{46.35} & \textbf{31.64} & \textbf{39.93} & \textbf{70.45} \\
PE ENC+DEC & 42.52 & 25.68 &33.70 & 59.93 \\
\end{tabular}
\caption{\label{test-results}
Metrics for models scored against the test-adapt set of unseen authors. 
}
\end{table}

\subsection{Author Performance Statistics}

\begin{table}[htpb]
\centering
\begin{tabular}{ccccccc}
\hline
\textbf{Model} & \textbf{min} & \textbf{max} & \textbf{mean} & \textbf{median} & \textbf{std} \\
\hline
HPI BASE & 12 & 41 & 27 & 29 & 29 \\
HPI ENC & 13 & 41 & 30 & 30 & 29 \\
HPI DEC & 13 & 43 & 31 & 32 & 29 \\
HPI ENC+DEC & 11 & 43 & 30 & 32 & 32 \\
\hline
AP BASE & 6 & 45 & 25 & 29 & 37 \\
AP ENC & 10 & 58 & 36 & 38 & 44 \\
AP DEC & 9 & 57 & 34 & 36 & 41 \\
AP ENC+DEC & 10 & 60 & 38 & 38 & 46 \\
\hline
PE BASE & 1 & 58 & 17 & 15 & 50 \\
PE ENC & 4 & 90 & 29 & 24 & 76 \\
PE DEC & 9 & 65 & 32 & 37 & 47 \\
PE ENC+DEC & 10 & 87 & 26 & 14 & 81 \\
\end{tabular}
\caption{ROUGE-2 statistics}\label{tab:stat}
\end{table}

In Table \ref{tab:stat} we report the statistics of the adapter authors performance on the unseen data. The min column is the lowest ROUGE-2 score from the 10 authors, the max column is the highest ROUGE-2 score from the 10 authors. With the exception of HPI ENC+DEC, all models improve upon the minimum ROUGE-2 score from that of the BASE model.

\subsection{Oracle Adaptation Model Performance}
We can observe the maximum achievable values from our adaptation technique by picking the best existing physician embedding based on looking at the entire test-adapt split, in order to investigate how much performance would increase if we picked the physician embedding that performs the best on the test-adapt split instead of the smaller adapt split. Due to space constraints we present only the ROUGE-2 scores via Table \ref{oracle-results}. In scenarios where a better physician embedding than the one chosen by the adaptation phase, the performance generally only improves by less than 1.0 ROUGE-2.

\begin{table}[htpb]
\centering
\begin{tabular}{ll}
\hline
\textbf{Oracle Model} & \textbf{R-2}\\
\hline
HPI BASE & 27.09  \\
HPI ENC  & 30.22  \\
HPI DEC & 31.74  \\
HPI ENC+DEC  & \textbf{31.92} \\
\hline
A\&P BASE  & 25.49  \\
A\&P ENC  & 36.57  \\
A\&P DEC  & 34.34  \\
A\&P ENC+DEC  & \textbf{38.34}  \\
\hline
PE BASE  & 16.72  \\
PE ENC  & 29.71  \\
PE DEC  & \textbf{32.80}  \\
PE ENC+DEC  & 26.09 \\
\end{tabular}
\caption{\label{oracle-results}
Metrics for Oracle adapted models scored against the test-adapt split of unseen authors. 
}
\end{table}

\section{Conclusion}
\label{sec:conclusion}

In this work, we introduced a novel method for rapid enrollment of new physicians by utilizing trained embeddings from previously enrolled physicians. We showed that through a novel adaptation algorithm, physician embeddings can be used with new physicians to improve performance without the need to retrain the model or embeddings.


\bibliographystyle{IEEEtran}
\bibliography{mybib}

\end{document}

%% file: keywords.tex
\keywords{summarization, embedding, personalization, transformer, doctor-patient-conversation}

%% file: abstract.tex
\begin{abstract}
    
    
    In this work, we present a novel technique to improve the quality of draft clinical notes for physicians. This technique is concentrated on the ability to model implicit physician conversation styles and note preferences. We also introduce a novel technique for the enrollment of new physicians when a limited number of clinical notes paired with conversations are available for that physician, without the need to re-train a model to support them. We show that our technique outperforms the baseline model by improving the ROUGE-2 score of the History of Present Illness section by 13.8\%, the Physical Examination section by 88.6\%, and the Assessment \& Plan section by 50.8\%.
    
\end{abstract}

